\documentclass[conference]{IEEEtran}
\IEEEoverridecommandlockouts
\usepackage{cite}
\usepackage{amsmath,amssymb,amsfonts}
\usepackage{algorithmic}
\usepackage{graphicx}
\usepackage{enumitem}
\usepackage{epstopdf}
\usepackage{textcomp}
\usepackage{xcolor}
\usepackage{algorithm}
\usepackage{multirow}
\DeclareMathOperator*{\argmin}{argmin}
\DeclareMathOperator*{\argmax}{argmax}
\def\BibTeX{{\rm B\kern-.05em{\sc i\kern-.025em b}\kern-.08em
    T\kern-.1667em\lower.7ex\hbox{E}\kern-.125emX}}
\begin{document}

\title{\textit{Auto-Model}: Utilizing Research Papers and HPO Techniques to Deal with the CASH problem}

\author{\IEEEauthorblockN{Chunnan Wang, Hongzhi Wang, Tianyu Mu, Jianzhong Li, Hong Gao}
\IEEEauthorblockA{\textit{Department of Computer Science} \\
\textit{Harbin Institute of Technology}\\
Harbin, China \\
\{WangChunnan, wangzh, mutianyu, lijzh, honggao\}@hit.edu.cn}
%\and
%\IEEEauthorblockN{Hongzhi Wang}
%\IEEEauthorblockA{\textit{Department of Computer Science} \\
%\textit{Harbin Institute of Technology}\\
%Harbin, China \\
%wangzh@hit.edu.cn}
%\and
%\IEEEauthorblockN{Tianyu Mu}
%\IEEEauthorblockA{\textit{Department of Computer Science} \\
%\textit{Harbin Institute of Technology}\\
%Harbin, China \\
%mutianyu@hit.edu.cn}
%\and
%\IEEEauthorblockN{Jianzhong Li}
%\IEEEauthorblockA{\textit{Department of Computer Science} \\
%\textit{Harbin Institute of Technology}\\
%Harbin, China \\
%lijzh@hit.edu.cn}
%\and
%\IEEEauthorblockN{Hong Gao}
%\IEEEauthorblockA{\textit{Department of Computer Science} \\
%\textit{Harbin Institute of Technology}\\
%Harbin, China \\
%honggao@hit.edu.cn}
}

\maketitle

\begin{abstract}
In many fields, a mass of algorithms with completely different hyperparameters have been developed to address the same type of problems. Choosing the algorithm and hyperparameter setting correctly can promote the overall performance greatly, but users often fail to do so due to the absence of knowledge. How to help users to effectively and quickly select the suitable algorithm and hyperparameter settings for the given task instance is an important research topic nowadays, which is known as the CASH problem. In this paper, we design the \textit{Auto-Model} approach, which makes full use of known information in the related research paper and introduces hyperparameter optimization techniques, to solve the CASH problem effectively. \textit{Auto-Model} tremendously reduces the cost of algorithm implementations and hyperparameter configuration space, and thus capable of dealing with the CASH problem efficiently and easily. To demonstrate the benefit of \textit{Auto-Model}, we compare it with classical Auto-Weka approach. The experimental results show that our proposed approach can provide superior results and achieves better performance in a short time.
\end{abstract}


\section{Introduction}

In many fields, such as machine learning, data mining, artificial intelligence and constraint satisfaction, a variety of algorithms and heuristics have been developed to address the same type of problem~\cite{p1,p2}. Each of these algorithms has its own advantages and disadvantages, and often they are complementary in the sense that one algorithm works well when others fail and vice versa~\cite{p2}. If we are capable of selecting the algorithm and hyperparameter setting best suited to the task instance, any particular task instance will be well solved, and our ability of dealing with the problem will be improved considerably \cite{p3}.

However, it is not trivial to achieve this goal. There are a mass of powerful and different algorithms to deal with a certain problem, and these algorithms have completely different hyperparameters, which have great effect on their performance. Even domain experts cannot easily and correctly select the appropriate algorithm with corresponding optimal hyperparameters from such a huge and complex choice space. Nonetheless, the suitable solution for the particular task instance is still desperately needed in practice. Therefore, the researchers presented combined algorithm selection and hyperparameter optimization (CASH) problem \cite{p4}, attempting to find easy approaches to help users simultaneously select the most suitable algorithm and hyperparameter setting to solve the practical task instance.

To the best of our knowledge, Auto-Weka \cite{p4} is the only approach that is capable of addressing this problem. Auto-Weka approach \cite{p4} transforms the CASH problem into a single hierarchical hyperparameter optimization problem, in which even the choice of algorithm itself is considered as a hyperparameter. Then it utilizes the effective and efficient hierarchical hyperparameter optimization technique \cite{p5,p6} to find the algorithm and hyperparameter settings appropriate to the given task instance. While Auto-Weka approach can deal with the CASH problem effectively, it causes two fatal shortcomings.

On the one hand, the algorithm implementation is quite complicated. Auto-Weka approach requires users or researchers to implement algorithms related to the problem before making a rational choice for the task instance. There are usually a mass of related algorithms, and generally a majority of them are not open source. If users want to solve the problem well utilizing Auto-Weka approach, a great deal of algorithms should be implemented, and this is extremely difficult and laborious. On the other hand, the configuration space is quite huge. The configuration space of hyperparameters of a single algorithm can be very large and complex \cite{p7}, let alone the configuration space, which considers the choice of algorithm and hyperparameters of many algorithms, in Auto-Weka approach. Searching the optimal configuration from such a huge space is very difficult, and this makes Auto-Weka unable to obtain good result within a short time.

We observe that many research papers related to machine learning have been proposed with a great deal of experiments, which carefully analyzed the performance of many related algorithms with certain hyperparameter settings on different task instances. Such reported experiences are pretty valuable to guide effective algorithm selection and reduce the search space. Thus, we attempt to adopt these experiences to deal with the CASH problem. However, the usage brings two challenges. One the one hand, it is nontrivial to extract the experiences in the research papers to the knowledge which could be used for the automatic algorithm selection. On the other hand, with the consideration that the existing knowledge may contain various kinds of algorithm (with different time complexity), the hyperparameter decision approach should be universal. However, existing approaches only apply to  some algorithms.

For the first challenge, we represent the machine learning task instances as a feature set, and model the knowledge as the mapping from the task instance to the optimal corresponding algorithm. Such mapping is constructed according to the experimental results reported in the research papers. With the consideration that different papers may report conflicting results and the experiences in papers are fragmented, we model the all the pieces of experiences as a information network, and resolve the conflicts and find such mapping with the information network. With the knowledge as experiences and the instances, we train a neural network to select the most suitable machine learning algorithm for the given task according to its features.
%而获取这样knowledge的方法是：利用关系网将有的experience拼接起来，利用网络图分析 挖掘出所有潜在关系并分析出task indtance的最优算法～
%given a machine learning task, the algorithm with the best performance could be retrieved. Thus, the algorithm is selected according to such information network.

For the second challenge, we combine Baysian and Genetic hyperparameter optimization (HPO) approach, which are complementary and cover almost all machine learning algorithm instances. For a given algorithm, we develop the strategy to determine whether Baysian or Genetic approach should be used according to the evaluation time on a small sample.

Major contributions of this paper are summarized as follows.

\begin{itemize}

\item We first propose to utilize the knowledge in research papers combining with HPO techniques to solve the CASH problem, and present \textit{Auto-Model} approach to deal with the CASH problem efficiently and easily. To the best of our knowledge, this is the first work to involve human experiences in algorithm selection and hyperparameter decision for data analysis.

\item We design the effective knowledge acquisition mechanism. The usable experience in the related papers are fragmented possible with conflict information. Our designed information integration approach and conflict resolve approach derives effective knowledge.

\item We design extensive experiments to verify the rationality of our \textit{Auto-Model} approach, and compare \textit{Auto-Model} with classical Auto-Weka approach. Experimental results show that the design of \textit{Auto-Model} is reasonable, and \textit{Auto-Model} has stronger ability of to deal with the CASH problem. It can provide a better result within a shorter time.
\end{itemize}

The remainder of this paper is organized into four sections. Section~\ref{section:2} discusses the HPO techniques used in our proposed approach, and defines some concepts related to HPO. Section~\ref{section:3} introduces our proposed \textit{Auto-Model} approach. Section~\ref{section:4} evaluates the validity and rationality of our proposed \textit{Auto-Model}, and compares \textit{Auto-Model} with classical Auto-Weka approach. Finally, we draw conclusions and discuss the future works in Section~\ref{section:5}.

\section{Prerequisites}\label{section:2}

In our proposed Auto-Model approach, the classical HPO techniques are used for some steps, including automatic feature identification, automatic neural architecture search and optimal hyperparameter setting acquisition. In this section, we introduce the HPO techniques used in \textit{Auto-Model}, and define some related concepts.

\subsection{HPO Techniques}\label{section:2.1}

Many modern algorithms, e.g., deep learning approaches and machine learning algorithms, are very sensitive to hyperparameters. Their performance depends more strongly than ever on the correct setting of many internal hyperparameters. In order to automatically find out suitable hyperparameter configurations, and thus promote the efficiency and effectiveness of the target algorithm, some HPO techniques~\cite{b1,b2,b3,b4,b5} have been proposed. Among them Grid Search (GS)~\cite{p11 }, Random Search (RS)~\cite{p12 }, Bayesian Optimization (BO)~\cite{p10} and Genetic Algorithm (GA)~\cite{p9} are very famous.

GS asks users to discretize the hyperparameter into a desired set of values to be studied, and then it evaluates the Cartesian product of these sets and finally chooses the best one as the optimal configuration. RS explores the entire configuration space, samples configurations at random until a certain budget for the search is exhausted, and outputs the best one as the final result. These two techniques have one thing in common, i.e. they ignore historical observations. That is, they fail to make full use of historical observations to intelligently infer more optimal configurations. This shortcoming often makes them incapable of providing the optimal solutions within short time, since the choice space they explore is always very complex and huge, and blind search can waste lots of time on useless configurations. BO and GA, which are used in our \textit{Auto-Model} approach, overcome this defect and exhibit better performance.

\textbf{BO} is a state-of-the-art optimization approach for the global optimization of expensive black box functions~\cite{p7}. It works by fitting a probabilistic surrogate model to all observations of the target black box function made so far, and then using the predictive distribution of the probabilistic model, to decide which point to evaluate next. Finally, consider the tested point with the highest score as the solution for the given HPO problem. Many works~\cite{p13,p14} apply BO to optimize hyperparameters of expensive black box functions due to its effectiveness.

\textbf{GA} is a heuristic global search strategy that mimics the process of genetics and natural selection. It works by encoding hyperparameters and initializing population, and then iteratively produces the next generation through selection, crossover and mutation steps. The iteration stops when one of the stopping criteria is met, and finally the optimal individual (i.e., configuration) is treated as the solution for the given HPO problem. GA is the intelligent exploitation of random search provided with historical data to direct the search into the region of better performance in the solution space. It is routinely used to generate high-quality solutions for complex optimization problems and search problems, due to its effectiveness.

Both BO and GA add intelligent analysis for better results. However, they are appropriate in different circumstances due to their different working principles. Each time BO infers an optimal configuration, it need take quite some time to estimate the posterior distribution of the target function using Bayesian theorem and all historical data. This working principle is suitable for the HPO problems whose tested algorithm has high complexity, and thus the hyperparameter configuration evaluations are very expensive and time-consuming (far more than BO's analysis time). The reason is that only few evaluations, which may be smaller than the size of population in GA, are allowed, and BO can make more thorough analysis of historical data and thus provide better solution.

As for GA, its analysis time (not include the time cost on configuration evaluations) is very short, and it can provide a totally new population, i.e., a large number of optimal configuration candidates, after analyzing each iteration. This working principle is suitable for the HPO problems whose tested algorithm has low complexity, and thus the hyperparameter configuration evaluations are cheap and fast (far less than BO's analysis time). The reason is that a large number of evaluations are allowed, and GA can fully bring into play the advantage of genetics and natural selection, and thus find out the excellent solution. In our \textit{Auto-Model} approach, we will choose to use GA or BO technique according to feature of the HPO problem.

\subsection{Concepts of HPO}\label{section:2.2}

Consider a HPO problem $P=(D,A,PN)$, where $D$ is a dataset, $A$ is an algorithm, and $PN=\{P_1,P_2,\ldots P_n\}$ are $n$ hyperparameters. We denote the domain of the $i^th$ hyperparameter $P_i$ by $\Lambda_{PN_i}$, and the overall hyperparameter configuration space of $PN$ as $\Lambda_{PN}=\Lambda{PN_1}\times \ldots \times \Lambda_{PN_n}$. We use $\lambda \in \Lambda_{PN}$ to represent a configuration of $PN$, and $f(\lambda,A,D)$ to represent the performance score of $A$ in $D$ under $\lambda$. Then, the target of the HPO problem $P=(D,A,PN)$ is to find

\begin{equation}
\lambda^\ast = \mathop{argmax} \limits_{\lambda \in \Lambda_{PN}} {f(\lambda,A,D)}
\end{equation}

from $\Lambda_{PN}$, which maximizes the performance of $A$ in $D$.

\section{\textit{Auto-Model} Approach}\label{section:3}

The target of \textit{Auto-Model} approach is to efficiently provide users with the high-quality solution for a task instance, including a quite appropriate algorithm and the optimal hyperparameter setting. To achieve this goal, we need to efficiently selected a suitable algorithm for the given task instance that users want to solve, and then efficiently find a proper hyperparameter setting for the selected algorithm. With many HPO approaches for various machine learning algorithms, the optimal setting search of our system is implemented by choosing a suitable and effective HPO technique. As for the algorithm selection, we propose to leverage the existing available information to obtain an effective decision-making model, which is used to make a good algorithm choice efficiently. We observe that research papers often report extensive performance experiments, which are pretty valuable to guide effective algorithm selection. Thus, we extract effective knowledge from these reported experiences to build the effective decision-making model to reduce manpower and resource consumption.

In Section~\ref{section:3.1}, we introduce some basic concepts on the knowledge in our approach. Section~\ref{section:3.2} gives the overall framework of \textit{Auto-Model}. Section~\ref{section:3.3} and Section~\ref{section:3.4} explain in detail the two main parts in \textit{Auto-Model}, respectively.

\begin{figure*}[t]
    \centering
    \includegraphics[width=1.0\textwidth]{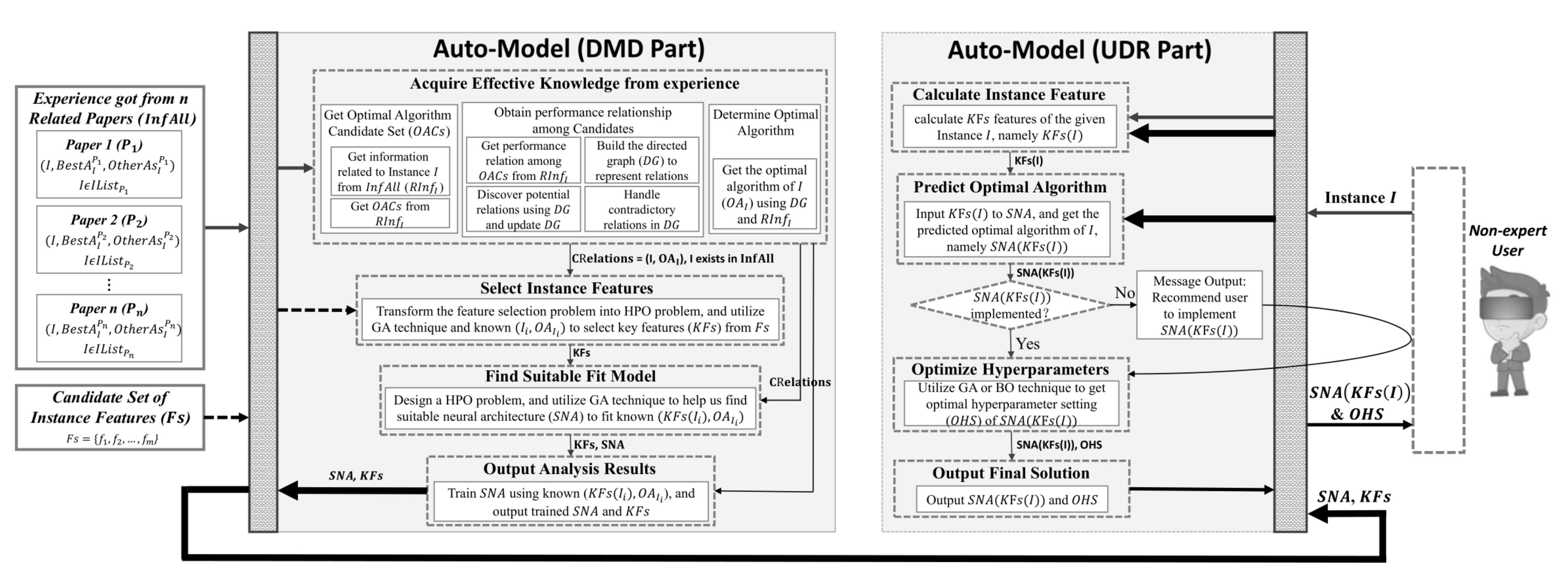}
    \caption{The overall framework of \textit{Auto-Model}. The DMD part of \textit{Auto-Model} (left) assists the UDR part of \textit{Auto-Model} (right) to make intelligent decisions. When the user inputs the task instance $I$ to solve, the UDR part of \textit{Auto-Model} provide the user with algorithm $SNA(KFs(I))$ and hyperparameter $OHS$ that are well suited to the given task instance $I$.}
\label{figure:1}
\end{figure*}

\subsection{Concepts}\label{section:3.1}

\textbf{Task Instance.} A task instance in machine learning corresponds to a dataset. For example, a task instance of the classification problem $I_C$ is a available dataset with category labels. A task instance could be described with a set of features called \emph{task instance features} (TIFs for brief) for the ease of algorithm selection with \textit{Auto-Model}. For different kind of task instances, TIFs may be different. Consider $I_C$, the features may include the number of records, numerical attribute and the predefined class number in $I_C$.

\textbf{Knowledge.} In the \textit{Auto-Model} approach, the extracted knowledge is used for providing guidance for the algorithm selection, which aims at selecting the most optimal algorithm ($OA_I$) for the given task instance ($I$). Therefore, the knowledge required in \textit{Auto-Model} is a set of pairs as the correspondence relationship between the task instance $I$ and its optimal algorithm $OA_{I}$, i.e. $(I,OA_{I})$.

\textbf{Experience.} Research papers may contain rich information. However, only a small share is useful for knowledge acquirement, which is called \emph{experience}. The algorithm with the highest performance on $I$ in each paper is a candidate of the $OA_I$. To further determine $OA_I$, the performance comparison relations among candidates are necessary. Thus, the experience required in \textit{Auto-Model} is a set of quadruples 
$(P,I,BestA_I^P,OtherAs_I^P)$, where $P$ is the paper that provide this piece of experience, $I\in IList_P$ is a task instance in $P_i$, $BestA^P_{I}$ is the algorithm with highest performance on $I$ in $P$, $OtherAs^P_I$ is the set of other algorithms analyzed in $P$ with lower performance than $BestA^P_{I}$, and $IList_P$ is the set of task instances analyzed in $P$.

%, and $d_i$ denotes the number of instances analyzed in $P_i$.}

The reason why we need $P$ is that there may exists conflict performance comparison relationships between two algorithms due to different experimental design or experimental errors. We can deal with these conflicts according to the reliability of papers, and thus get more reliable performance relationship, as will be discussed in Section~\ref{section:3.3.1}.

\subsection{Overall Framework}\label{section:3.2}

Fig.~\ref{figure:1} gives the overall framework of our proposed \textit{Auto-Model} with two major components: Decision-Making Model Designer (DMD) and User Demand Responser (UDR). DMD (introduced in Section~\ref{section:3.3}) selects and trains the suitable model for the algorithm selection, which contains three steps.
%takes the knowledge obtained from related papers $InfAll$ and the candidate set of TIF $Fs$ as input. Its target is to utilizes these available relevant information to obtain an effective decision-making model that can accurately map task instance features to the a quite suitable algorithm. Before finding the most suitable decision-making model from all candidats, two works should be completed.

The first step acquires knowledge from the paper set (introduced in Section~\ref{section:3.3.1}). The second step selects suitable features from feature candidates $Fs$ to represent the task instance (introduced in Section~\ref{section:3.3.2}), which is taken by the model as the input. Then in the third step, the effective model is selected and trained based on the knowledge from step 1 and the features from step 2 (introduced in Section~\ref{section:3.3.3}).

The UDR (introduced in Section~\ref{section:3.4}) takes the well-trained decision-making model $SNA$ whose input contents are $Fs$, and the task instance $I$ as the input. It interacts with the users, and aims at responding reasonably rapidly to the user demand and providing users with the high-quality solution by making the best of the suitable HPO technique and $SNA$. $SNA$ can help UDR to quickly select a suitable algorithm from large amount of choices, and thus tremendously reduce the search space. And the selected suitable HPO technique can quickly promote the performance of the selected algorithm. Their cooperation makes UDR capable of providing high-quality solution within shorter time.

%\textcolor{blue}{The DMM part provides strong support for the intelligent decision of the RUD part, and the RUD part offers high quality service to users. Two major parts of \textit{Auto-Model} cooperate closely and thus be able to solve the CASH problem effectively and efficiently.}

\begin{figure*}[t]
    \centering
    \includegraphics[width=1.0\textwidth]{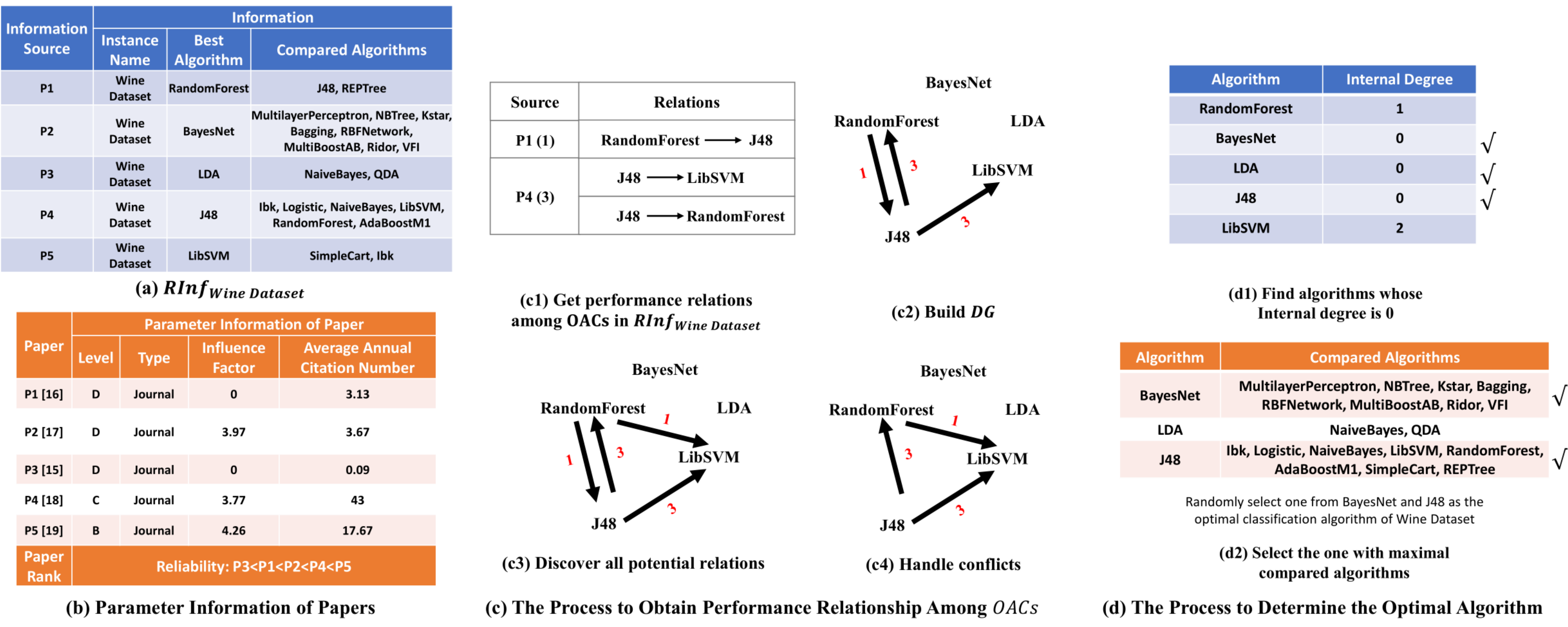}
    \caption{An example of the process to acquire a piece of knowledge, i.e., the correspondence between an task instance (Wine Dataset) and its optimal classification algorithm. Suppose $RInf_{Wine Dataset}$ is shown in (a), and the parameter information of papers involved \cite{G3,G4,G11,G15,G18} are in (b). Then, $OACs=\{RandomForest, BayesNet, LDA, J48, LibSVM\}$, the process to obtain performance relationship among $OACs$ are shown in (c), and (d) is the process that we determine the optimal classification algorithm (BayesNet or J48) of Wine Dataset. Note that we only consider the algorithms implemented in Weka or Sklearn library of Python in this example.}
\label{figure2}
\end{figure*}

\subsection{Decision-Making Model Designer (DMD)}\label{section:3.3}

\subsubsection{Knowledge Acquiremet}\label{section:3.3.1}

Whether we want to select instance features or find the suitable fit model, the knowledge that describes the correspondence between the task instance and its optimal algorithm is necessary, since it is the basis for the rationality evaluation of the feature set and decision-making model.

%Therefore, the first step in the DMM part of \textit{Auto-Model} is to acquire effective knowledge from the experience.

%\textbf{Motivation.} Consider the knowledge model
%
%
%Suppose papers are users, algorithms are products, and task instances are application scenarios, then, a piece of experience $(P_i,$$I_i^{j},$$BestA_i^{j},$$OtherAs_i^{j})$\footnote{$P_i$ denotes a related paper, $I_i^{j}$ is the name of the $j^{st}$ instance in $P_i$, $BestA_i^{j}$ is the algorithm with highest performance on $I_i^{j}$ in $P_i$, $OtherAs_i^{j}$ are the algorithms which are proved to have worse performance than $BestA_i^{j}$ on $I_i^{j}$ in $P_i$ (j=1,...,$d_i$). $d_i$ denotes the number of instances analyzed in $P_i$.} can be treated as a user's feedback on the performance of certain products in an application scenario\footnote{Each user has a different confidence level and has his own way of evaluating products.}. In the reality, people often introduce trust evaluation mechanism and put the fragmentation information (i.e., known feedbacks) together according to credibility and internal relations, and finally get the best product in different application scenarios by analyzing the complete information network and recommend best products to more users. We can borrow this idea to effectively deal with our knowledge acquisition problem.

%\textbf{Design Idea.}

The key points of effective knowledge extraction are complete information network building, and to design our own judgment standards of the optimal algorithm. Let $InfAll$ denote all usable experience extracted from related papers, and $RInf_I$ be the experience related to instance $I$ in $InfAll$. Then, the best algorithm of $I$ should be among $OACs$=$\{$BestAs contained in $RInf_I$$\}$. However, to judge which one is the best, we need as many performance relations among $OACs$ as possible for assistance. Therefore, in our knowledge acquisition problem, the complete information network is a directed graph $DGraph$ that contains all potential performance relationships among $OACs$.

$RInf_I$ provides us with some performance relations. Considering a tuple $(P,$$I,$$A_i,$$OtherAs)$ in $RInf_I$, if there exists $A_j$$\in$$OtherAs$ satisfying $A_j$$\in$$OACs$, then we add a directed edge $A_i$$\rightarrow$$A_j$ with weight $Rel_{ij}$ (the reliability value of paper $P$). We can also apply the breadth-first search on each algorithm in $DGraph$ to obtain other potential relationships among $OACs$. Now, we obtain all available performance relationships among $OACs$.

Note that there may exists contradictory relations in $DGraph$, due to the different experimental designs of different papers or the experimental errors of certain papers. We propose to use the reliability of the relations, i.e., edge weight, to handle these conflicts. We only preserve one directed edge with the highest weight. Now, we obtain a reasonable and complete information network $DGraph$ related to $I$. We can acquire the optimal algorithm of $I$ by analyzing $DGraph$.

The algorithm whose in-degree is $0$ in $DGraph$ is proved to have better performance on $I$, and we can consider it as the optimal algorithm of $I$, denoted by $OA_I$. However, more than one candidates in $DGraph$ may satisfy this condition, due to the inadequacy of the available relations. In this situation, we propose to analyze the comparison experience of each candidate, i.e., the number of algorithms that are proved to be less effective than the candidate according to $RInf_I$ and $DGraph$. And we select the one with the richest experience as the $OA_I$. Thus, we obtain a piece of knowledge $(I,$$OA_I)$, and acquire many such knowledge from $InfAll$ in this way. Fig.~\ref{figure2} is an example of the process to acquire an piece of knowledge.

\textbf{Detail Workflow.} Algorithm~\ref{alg:CA} shows the pseudo code of knowledge acquisition approach. Firstly, it collects all instances in $InfAll$ ($IList$) and the reliability value of each paper involved in $InfAll$ (the index of the paper in $PRank$), and initializes $CRelations$ (Line 1-3). Then, the iteration begins, for each instance $I$ in $IList$, KnowledgeAcquisition follows the process mentioned above to acquire its optimal algorithm (Line 5-15).  The details are as follows. The information related to $I$ ($RInf_I$) and the optimal algorithm candidate set of $I$ ($OACs$) are obtained first (Line 5-7). Then the performance relations among $OACs$ in $RInf_I$ are extracted (Line 8), and $DGraph$ representing the performance relations among $OACs$ is built (Line 9). After that breadth-first search is applied and all potential relations are discovered and added to $DGraph$ (Line 10-11), and the contradictory relations in $DGraph$ are handled (Line 12). Now, $DGraph$ contains all available and reasonable relations among $OACs$, and the optimal algorithm of $I$ ($OA_I$) is identified with the help of  $DGraph$ and $RInf_I$ (Line 13-15). In this way, a piece knowledge $(I,$$OA_I)$ is acquired. Note that in order to improve the reliability of the acquired knowledge, we do not consider the knowledge related to the instance $I$ in $IList$, whose $RInf_I$ contains very few algorithms (Line 6). The reason is that in this situation, insufficient performance comparisons are involved in $RInf_I$, and $OA_I$ lacks of sufficient evidences to be explained. We collect knowledge with sufficient evidences and finally get the result $CRelations$ (Line 16-19).

\begin{table}[t]
\newcommand{\tabincell}[2]{\begin{tabular}{@{}#1@{}}#2\end{tabular}}
\caption{The bases for the comparison of paper reliability.}\smallskip
\centering
\resizebox{\columnwidth}{!}{
\smallskip\begin{tabular}{|l|r|r|r|r|}
\hline
\textbf{Paper Parameter} & \textbf{Priority Level} & \textbf{Parameter Type} & \textbf{Ranges or Options} & \textbf{\tabincell{c}{Reliability Comparison \\Strategy}} \\
\hline
Paper level & 1 & list & A, B, C, D & A$>$B$>$C$>$D \\
Paper type & 2 & list & Journal, Conference & Journal$>$Conference \\
Influence factor & 3 & float & $\geq$0 & The bigger the better \\
\tabincell{c}{Average annual\\citation number} & 4 & int & $\geq$0 & The bigger the better \\
\hline
\end{tabular}
}
\label{table1}
\end{table}

\subsubsection{Instance Features Selection}\label{section:3.3.2}

In our \textit{Auto-Model} approach, we select the suitable algorithm for the given task instance according to its features.  An instance may have many possible features, but not all of them are correlated to the algorithm performance. Selecting features correlated to the algorithm performance to represent the instance can not only reduce the feature calculation cost, but also help algorithm selection approach to better differentiate between instances and thus be more effective. Because of these benefits, we design the algorithm to automatically select suitable task instance features from the candidate feature set denoted by $Fs$.

\begin{algorithm}[t]
\caption{KnowledgeAcquisition Approach}
\label{alg:CA}
\begin{algorithmic}[1]
\REQUIRE
Experience obtained from n related papers \\$InfAll$=$\{(P_i,$$I,$$BestA_I^{P_i},$$OtherAs_I^{P_i})$$|$$i$=$1$,...,$n$,$I$$\in$$IList_P$$\}$
\ENSURE Some effective knowledge $CRelations$

\STATE $IList$ $\gets$  all instances involved in $InfAll$
\STATE $PRank$ $\gets$ rank papers in $InfAll$ in ascending order of their reliability according to strategies in TABLE~\ref{table1}
\STATE $CRelations$ $\gets$ $\emptyset$
\FOR{$I$ $\in$ $IList$}
	\STATE $RInf_I$ $\gets$ tuples related to instance $I$ in $InfAll$
	\IF{$RInf_I$ involves $>5$ algorithms}
		\STATE $OACs$ $\gets$ $\{ t[2] | t \in RInf_I \}$
		\STATE $Relations$ $\gets$ $\{$ $(A_i,$$A_j,$$Rel_{ij})$ $|$ $A_i$,$A_j$$\in$$OACs,$ $Base_{ij}$$\neq$$\emptyset$, $Rel_{ij}$=max value in $Base_{ij}$ $\}$ \# $Base_{ij}$= $\{$$PRank$.index(t[0])$|$t$\in$$RInf_I$\&t[2]=$A_i$\&$A_j$$\in$t[3]$\}$
		\STATE $DGraph$ $\gets$ build directed graph according to $Relations$, where $(A_i,A_j,Rel_{ij})$ $\in$ $Relations$ denotes a directed edge $A_i$ $\rightarrow$ $A_j$ with weight $Rel_{ij}$
		\STATE for each node $A_i$ in $DGraph$, start from it and apply breadth-first search. Record all nodes visited ($BFSA_i$), and the minimum weight in the path from $A_i$ to $A_j\in BFSA_i$ ($NRel_{ij}$)
		\STATE $DGraph$ $\gets$ $\{$ $(A_i,$$A_j,$$NRel_{ij})$ $|$ $A_i$$\in$$OACs$,  $A_j$$\in$$BFSA_i$ $\}$ \# update $DGraph$
		\STATE $DGraph$ $\gets$ if 2 nodes $A_i$,$A_j$ in $DGraph$ have conflict relations, only preserve one with bigger weight
		\STATE $OACs$ $\gets$ nodes in $DGraph$ with no internal edges
		\STATE $OACs$ $\gets$ $\{$ $(A_i,$$|ComAs_i|)$ $|$ $A_i$$\in$$OACs,$ $ComAs_i$=$\{$t[3]$|$t$\in$$RInf_I$\&t[2]$\in$$BFSA_i\}$ $\}$
		\STATE $OA_I$ $\gets$ an algorithm in $OACs$ with highest score
		\STATE $CRelations$ $\gets$  $CRelations$ $\bigcup$ $\{(I,OA_I)\}$
	\ENDIF
\ENDFOR
\RETURN $CRelations$
\end{algorithmic}
\end{algorithm}

\textbf{Motivation.} To select a suitable feature subset from $Fs$, we should define a metric $M$ to reasonably evaluate the quality of the selected feature subsets. Since the available information in this step is $Fs$ and the obtain knowledge as the correspondence relations $CRelations$=$\{$$(I_i,$$OA_{I_i})$$|$$i$=1,...,t$\}$, which could be treated as a classification dataset, we have to find a method that utilizes such information to compute $M$.

It is known that when unrelated features are involved or correlated features are not completely considered, the performance of the classification model will be greatly affected, since much noise will cause much interference, and lacking of important features will make it hard to differentiate some records with different categories. This fact makes it feasible to utilize the known classification dataset to obtain $M$. We can select a classification model $CM$, e.g., a MLP classifier. And for each feature subset $FSub$, we use the performance score of $CM$ on the classification sub-dataset $\{$$(FSub(I_i),$$OA_{I_i})$$|$$i$=1,...,t$\}$ to assess the quality of $FSub$. The higher the score is, the better $FSub$ is. Thus, we get $M$, and can find suitable instance features with the help of $M$.

\textbf{Design Idea.} According to above discussions, the problem of finding the feature subset with the highest score is transformed into a HPO problem $P$=$(D,$$A$,$PN)$ aiming at finding the optimal configuration of $PN$ that maximizes the performance of $A$ in $D$. In this problem, we consider $\{$$(Fs(I_i),$$OA_{I_i})$$|$i$=1,...,t\}$ as $D$, a multilayer perceptron (MLP) classifier with default structure as $A$, and the features in $Fs$ as hyperparameters $PN$. Each feature $f_i$ corresponds to a hyperparameter with two options, i.e, ``True'' meaning ``consider $f_i$ in $D$'' and ``False'' meaning ``ignore $f_i$ in $D$''. Thus, we convert the instance feature selection problem to a HPO problem $P$=$(D,$$A,$$PN)$. We can utilize the classical HPO algorithm to deal with $P$ effectively, and finally obtain suitable instance features according to the optimal configuration of $PN$ provided by the HPO technique. In Section~\ref{section:2}, we have pointed out that two classical and well-performed HPO techniques, i.e., BO and GA, are suit for different circumstances. Due to the fact that there are not many instances in the related research papers (generally less than $10^3$), the dataset $D$ in $P$ is small, and the hyperparameter configuration evaluations in $P$ are pretty fast and cheap. Such situation is suitable for GA. As the result, we choose to use GA to deal with $P$ designed in this part.

\begin{algorithm}[t]
\caption{FeatureSelection Approach}
\label{alg:FS}
\begin{algorithmic}[1]
\REQUIRE Known knowledge $CRelations$ = $\{(I_{1},OA_{I_{i}}),\ldots , (I_{t},OA_{I_{t}})\}$, and candidate set of instance features $Fs$ = $\{f_{1},\ldots , f_{m}\}$
\ENSURE Key features $KFs$

\STATE $D$ $\gets$  $\{$ $(Fs(I_{i}),$$OA_{I_{i}})$ $|$ $(I_{i},$$OA_{I_{i}})$$\in$$CRelations$ $\}$ \\ \# $Fs(I_{i})$ represents the feature vector of instance $I_{i}$, which contains all features in $Fs$
\STATE $PN$ $\gets$ for each feature $f_{i}$$\in$$Fs$, construct a boolean hyperparameter $f_{i}$, where True' (`False') means consider (ignore) feature $f_{i}$ in the given dataset $D$.
\STATE $A$ $\gets$ a MLP classifier with default architecture and parameter setting
\STATE construct a HPO problem $P$ = $(D,A,PN)$ \\\# The k-fold cross-validation accuracy is used to calculate f($\lambda$$\in$$\Lambda PN$,A,D)
\STATE $OptimalConf$ $\gets$ GA($P$) (group size: 50, evolutional epochs: 100)
\STATE $KFs$ $\gets$ $\{$ $f_{i}$ $|$ $f_{i}$$\in$$Fs$\&$OptimalConf[f_{i}]$=`True' $\}$
\RETURN $KFs$
\end{algorithmic}
\end{algorithm}

\textbf{Detail Workflow.} Algorithm~\ref{alg:FS} shows the pseudo code of instance feature selection approach. Firstly, FeatureSelection approach designs a HPO problem $P$=$(D,$$A,$$PN)$ related to the instance feature selection (Line 1-4). Then, it applies the GA technique to deal with $P$ and obtain an optimal configuration of $PN$ ($OptimalConf$) (Line 5). Finally, it obtains key features in $Fs$ by picking out features that are set to ``True'' in $OptimalConf$ (Line 6-7).

\subsubsection{Model Training}\label{section:3.3.3}

Based on the key instance features $KFs$ and knowledge $CRelations$=$\{$$(I_i,$$OA_{I_i})$$|$$i$=1,...,t$\}$, DMD trains the decision-making model that accurately maps $KFs(I)$ to $OA_I$, so as to help UDR make reasonable decisions.

\textbf{Motivation.} The difficulty is to ensure the precision of the model. The ability of most classification algorithms and regression algorithms to deal with the new dataset related to $KFs(I)$ and $OA_I$ are unsure, since there has not been a theory or a study yet to explain clearly their ability to deal with different datasets, to the best of our knowledge. If we select a model from such kind of algorithms, there is a very good chance that no high-precision models will be found in the end. Therefore, we do not consider this kind of algorithms. Since Neural networks are proved to be capable of approximating any function by arbitrary precision in theory \cite{p8}, we choose to use the multilayer perception (MLP), a feedforward artificial neural network, as our fit model.

Note that the architecture of MLP has great effect on its performance. Therefore, to achieve high precision, we need to design a proper architecture for MLP. We can utilize known $(KFs(I)$,$OA_I)$ pairs to evaluate the quality of the architecture of MLP, and thus find a high-precision fit model under the guidance of the quality score.

\begin{algorithm}[t]
\caption{ArchitectureSearch Approach}
\label{alg:AS}
\begin{algorithmic}[1]
\REQUIRE Known knowledge $CRelations$ = $\{(I_{1},OA_{I_{i}}),\ldots , (I_{t},OA_{I_{t}})\}$, key features $KFs$, and $Precision$
\ENSURE Suitable neural architecture $SNA$

\STATE $D$ $\gets$ $\{$ $(KFs(I_{i}),$$OneHot^{'}(OA_{I_{i}}))$ $|$ $(I_{i},$$OA_{I_{i}})$ $\in$ $CRelations$ $\}$
\STATE $PN$ $\gets$ hyperparameters of MLP (shown in TABLE~\ref{table2})
\STATE $A$ $\gets$ a MLP regressor
\STATE construct a HPO problem $P$ = $(D,A,PN)$ \# The k-fold cross-validation MSE (mean squared error) is used to calculate $f($$\lambda$$\in$$\Lambda PN,$$A,$$D)$
\STATE $OptimalConf$ $\gets$ $GA(P)$ (group size: 50) \# GA stops when $\lambda$$\in$$\Lambda PN$ whose $f($$\lambda,$$A,$$D)$$<$$Precision$ is found
\STATE $SNA$ $\gets$ an MLP regressor with $OptimalConf$ setting
\RETURN $SNA$
\end{algorithmic}
\end{algorithm}

\begin{table}[t]
\newcommand{\tabincell}[2]{\begin{tabular}{@{}#1@{}}#2\end{tabular}}
\caption{Ten Hyperparameters of MLP.}\smallskip
\centering
\resizebox{\columnwidth}{!}{
\smallskip\begin{tabular}{|l|c|c|l|}
\hline
\textbf{Name} & \textbf{Type} & \textbf{Set ranges or available options} & \textbf{Meaning} \\
\hline
hidden layer & int & 1-20 & The number of hidden layer in MLP \\
hidden layer size & int & 5-100 & \tabincell{l}{The number of neuron in each hidden layer} \\
activation & list & [`relu',`tanh',`logistic',`identity'] & The activation used on each neuron \\
solver & list & [`lbfgs',`sgd', `adam'] & The solver used to optimize MLP \\
learning rate & list & [`constant',`invscaling',`adaptive'] & \tabincell{l}{Used for weight updates, only used when\\ solver is `sgd'}\\
max iter & int & 100-500 & Maximum number of iterations \\
momentum & float & 0.01-0.99 & \tabincell{l}{Momentum gradient descent update, only\\ used when solver is `sgd'} \\
validation fraction & float & 0.01-0.99 & \tabincell{l}{Proportion of reserved training sets for early\\ morning stop validation} \\
beta 1 & float & 0.01-0.99 & \tabincell{l}{The exponential decay rate of the estimation\\ of the first order moment vector} \\
beta 2 & float & 0.01-0.99 & \tabincell{l}{The exponential decay rate of the estimation\\ of the second order moment vector} \\
\hline
\end{tabular}
}
\label{table2}
\end{table}

\textbf{Design Idea.} The problem of finding the proper MLP architecture with the highest score can also be transformed into a HPO problem. Consider $\{$$(KFs(I_i),$OneHot'($OA_{I_i}$)$)$$|$$i$=1,...,t$\}$ as $D$\footnote{To obtain OneHot'($OA_{I_i}$), firstly change $OA_{I_i}$ into the one hot label OneHot($OA_{I_i}$), where except for the index correspond to $OA_{I_i}$ all other places are 0, then set the position of algorithms which cannot deal with ithe nstance $I_i$ (e.g., some classification algorithms cannot deal with the instances with neural features) into -1.}, a MLP regressor as $A$, and consider hyperparameters in TABLE~\ref{table2}, which decide the architecture of MLP, as $PN$. Thus, we convert the MLP architecture search problem to a HPO problem $P$=$(D,$$A,$$PN)$.

We can utilize the classical HPO algorithm to deal with $P$ effectively, and finally obtain a proper architecture according to the optimal configuration of $PN$ provided by the HPO algorithm. Note that, to avoid selecting algorithms that are unable to deal with the given instance, we use OneHot'($OA_{I}$) instead of $OA_I$ or OneHot($OA_I$) as the output of MLP, and we choose to use MLP regressor instead of classifier because of this output format. Besides, note that the dataset $D$ in $P$ is small, and the hyperparameter configuration evaluations in $P$ are fast and cheap. Therefore, we choose to use GA to deal with $P$ designed in this part.

\begin{algorithm}[t]
\caption{AutoModelDMD Approach}
\label{alg:AMOff}
\begin{algorithmic}[1]
\REQUIRE Experience obtained from n related papers \\$InfAll$=$\{(P_i,$$I,$$BestA_I^{P_i},$$OtherAs_I^{P_i})$$|$$i$=$1$,...,$n$,$I$$\in$$IList_P$$\}$, and candidate set of instance features $Fs$ = $\{f_{1},\ldots , f_{m}\}$
\ENSURE Key features $KFs$, and suitable neural architecture $SNA$

\STATE $CRelations$ $\gets$ CorrespondenceAcquisition($InfAll$)
\STATE $KFs$ $\gets$ FeatureSelection($CRelations$,$Fs$)
\STATE $SNA$ $\gets$ ArchitectureSearch($CRelations$,$KFs$,$Precision$ = -0.0015) \# we set $Precision$ to -0.0015 by default
\STATE $D$ $\gets$ $\{$ $(KFs(I_{i}),$$OneHot^{'}(OA_{I_{i}}))$ $|$ $(I_{i},$$OA_{I_{i}})$ $\in$ $CRelations$ $\}$
\STATE $SNA$ $\gets$ train MLP regressor with $SNA$ setting using $D$
\RETURN $KFs$, $SNA$
\end{algorithmic}
\end{algorithm}

\textbf{Detail Workflow.} Algorithm~\ref{alg:AS} shows the pseudo code of MLP architecture search approach. Firstly, the algorithm constructs a HPO problem $P$=$(D,$$A,$$PN)$ according to the MLP architecture search (Line 1-4). Then, it applies the GA algorithm to deal with $P$ and obtain an optimal configuration of $PN$ ($OptimalConf$) which makes the precision of MLP high (Line 5). Finally, it obtains a MLP architecture according to $OptimalConf$ (Line 6-7).

\textbf{Complexity Analysis.} Combining the three steps organically, then we obtain global picture of DMD, which is shown in Algorithm~\ref{alg:AMOff}. The KnowledgeAcquisition mainly analyzes $InfAll$ with time complexity $O(p^2)$, where $p$ is the number of tuples in $InfAll$. As for the FeatureSelection and ArchitectureSearch, computing the features of the instances in $InfAll$ and running the GA algorithm dominate the time, their time complexity is $O(pm+g)$, where $m$ is the number of features in $Fs$, and $g$ is the number of generations used in ArchitectureSearch. In all, the time complexity of DMD is $O(p^2+pm+g)$.

\subsection{User Demand Responser (UDR)}\label{section:3.4}

The goal of UDR, is to efficiently provide users with effective solution, including the suitable algorithm and its optimal hyperparameter setting. If UDR searches the optimal solution from a huge search space, which contains the related algorithms, the cost will be pretty large. Therefore, its first step is to prune the search space by determining a quite suitable algorithm utilizing the effective decision-making model obtained by DMD. Then, it only considers the selected algorithm and choose a suitable HPO technique to optimize its hyperparemeters to improve the performance. In Section~\ref{section:2}, we have analyzed that BO and GA suit for different algorithms. Selecting a suitable HPO technique according to the algorithm feature discovered with a small sample can get better hyperparameter setting within short time. In this way, UDR obtains a high-quality solution.

\textbf{Detail Workflow.} Algorithm~\ref{alg:AMOn} gives the pseudo code of UDR. UDR of \textit{Auto-Model} takes: (1) an task instance $I$ which is provided by users, (2) key features $KFs$ and a trained MLP with a suitable architecture $SNA$ which are the findings of the DMD, as input. It determines a suitable algorithm ($SA$) for $I$ with the help of $KFs$ and $SNA$ (Line 1). Then, it automatically finds the optimal hyperparameter setting ($OHS$) of the chosen algorithm by making full use a suitable HPO technique (Line 2-4). Finally, it provides users a reasonable solution ($SA$,$OHS$) (Line 5).

\textbf{Complexity Analysis.} Calculating the key features of the given instance $I$ and running the HPO algorithm dominate the running time of UDR. The time complexity of calculating instance features is $O(kd^2)$, where $d$ is the dimension of the input instance, and the time cost by HPO techniques is determined by the users.

\section{Experiments}\label{section:4}

In the experiments, we test the proposed approach on classification CASH problem, which aims at finding the most suitable classification algorithm with the optimal hyperparameter setting in Weka~\footnote{In the experiments, various classification algorithms should be implemented for examining the CASH techniques. To ensure the fairness of the comparison, we adopt the implementation of the classification algorithms in Weka, an open source software, which contains large amount of classification algorithms. We simplify the problem by only considering the classification algorithms implemented in Weka, and utilize the CASH-Weka problem to examine CASH techniques.} for the given classification dataset (we then denote this CASH problem by CASH-Weka). We use the CASH-Weka problem to explain the rationality of our proposed \textit{Auto-Model} approach (Section~\ref{section:4.1}), and compare the effectiveness of \textit{Auto-Model} and Auto-Weka approach (Section~\ref{section:4.2}). We implement all the approaches in Python, and run experiments on a machine with an Intel 2.3GHz i5-7360U CPU and 16GB of memory.

\begin{algorithm}[t]
\caption{AutoModelUDR Approach}
\label{alg:AMOn}
\begin{algorithmic}[1]
\REQUIRE An instance $I$, key features $KFs$, and the suitable neural architecture $SNA$
\ENSURE An optimal  algorithm $SA$ and its optimal hyperparameter setting $OHS$

\STATE $SA$ $\gets$ $SNA(KFs(I))$ \# If $SA$ has not been implemented yet, notify the user to implement it
\STATE $PN$ $\gets$ the hyperparameters of $SA$
\STATE construct a HPO problem $P$ = $(I,SA,PN)$
\STATE $OHS$ $\gets$ HPOAlg($P$) \\ \# HPOAlg is BO or GA. If the calculation of $f(\lambda,$$SA,$$I)$ generally costs less than 10 minutes, then we set HPOAlg=GA, else, HPOAlg=BO\\ \# User can stop HPOAlg at any time, and $OHS$ is the optimal configuration obtained so far
\RETURN $SA$, $OHS$
\end{algorithmic}
\end{algorithm}

\begin{table*}[t]
\newcommand{\tabincell}[2]{\begin{tabular}{@{}#1@{}}#2\end{tabular}}
\caption{The classification dataset features. Suppose $D$ is a classification dataset with $m$ records, $n$ common attributes $\{A_{1}$,$\ldots$,$A_{n}\}$ and a target attribute $AT$. We use $ANList$ to represent all numeral attributes in $D$, and $ACList$ to represent all categorical attributes in $D$. For an attribute $A_{i}$$\in$$ANList$, we use $Var(A_{i})$ to denote the variance of the $A_{i}$ values in $D$, and $Avg(A_{i})$ to denote the average value of the $A_{i}$ values in $D$. For an attribute $A_{i}$$\in$$ACList$, we use $A_{i}[n]$ to denote the number of classes of $A_{i}$ in $D$, $A_{i}[c_{j}]$ (j=1,...,$A_{i}[n]$) to denote all classes of $A_{i}$, and $Num(A_{i}[c_{j}])$ to denote the number of records whose $A_{i}$ is $A_{i}[c_{j}]$ in $D$.}\smallskip
\centering
\resizebox{0.9\textwidth}{!}{
\smallskip\begin{tabular}{|c|c|l|}
\hline
\textbf{Symbol} & \textbf{Formula} & \textbf{Meaning} \\
\hline
$f_1$ & $AT[n]$ & The number of classes in the target attribute \\
$f_2$ & $H(AT)$ & The entropy of the classes in the target attribute \\
$f_3$ & $\max\limits_{1\leq j\leq AT[n]}$$\frac{Num(AT[c_j])}{m}$ & \tabincell{l}{The proportion of the class, which accounts for the highest proportion in the target attribute} \\
$f_4$ & $\min\limits_{1\leq j\leq AT[n]}$$\frac{Num(AT[c_j])}{m}$ & \tabincell{l}{The proportion of the class, which accounts for the lowest proportion in the target attribute} \\
$f_5$ & $|ANList|$ & The number of numeral attributes in the dataset \\
$f_6$ & $|ACList|$ & The number of categorical attributes in the dataset \\
$f_7$ & $\frac{|ANList|}{n}$ & The proportion of numeral attributes in all common attributes \\
$f_8$ & $n$ & The number of common attributes \\
$f_9$ & $m$ & The number of records \\
$f_{10}$ &	$\min\limits_{A_i\in ACList\& A_i\neq AT}$$A_i[n]$ & \tabincell{l}{The number of classes of a common and categorical attribute, which has the fewest classes} \\
$f_{11}$ & $H(A^{\#})$ & \tabincell{l}{The entropy of a common and categorical attribute, which has the fewest classes\\ (The meaning of $A^{\#}$=$\argmin\limits_{A_i\in ACList\& A_i\neq AT}$$A_i[n]$ is the same in $f_11$, $f_12$ and $f_13$)} \\
$f_{12}$ & $\max\limits_{1\leq j \leq A^{\#}[n]}$$\frac{Num(A^{\#}[c_j])}{m}$ & \tabincell{l}{The proportion of the class, which accounts for the highest proportion in a common and\\ categorical attribute, that has the fewest classes} \\
$f_{13}$ & $\min\limits_{1\leq j\leq A^{\#}[n]}$$\frac{Num(A^{\#}[c_j])}{m}$ & \tabincell{l}{The proportion of the class, which accounts for the lowest proportion in a common and\\ categorical attribute, that has the fewest classes} \\
$f_{14}$ & $\max\limits_{A_i\in ACList\& A_i\neq AT}$$A_i[n]$ & \tabincell{l}{The number of classes of a common and categorical attribute, which has the most classes} \\
$f_{15}$ & $H(A^{\star})$ & \tabincell{l}{The entropy of a common and categorical attribute, which has the most classes\\ (The meaning of $A^{\star}$=$\argmax\limits_{A_i\in ACList\& A_i\neq AT}$$A_i[n]$ is the same in $f_15$, $f_16$ and $f_17$)} \\
$f_{16}$ & $\max\limits_{1\leq j\leq A^{\star}[n]}$$\frac{Num(A^{\star}[c_j])}{m}$ & \tabincell{l}{The proportion of the class, which accounts for the highest proportion in a common and\\ categorical attribute, that has the most classes} \\
$f_{17}$ & $\min\limits_{1\leq j\leq A^{\star}[n]}$$\frac{Num(A^{\star}[c_j])}{m}$ & \tabincell{l}{The proportion of the class, which accounts for the lowest proportion in a common and\\ categorical attribute, that has the most classes} \\
$f_{18}$ & $\min\limits_{A_i\in ANList}$$Avg(A_i)$ & The minimum value in the average values of numeral attributes \\
$f_{19}$ & $\max\limits_{A_i\in ANList}$$Avg(A_i)$ & The maximum value in the average values of numeral attributes \\
$f_{20}$ & $\min\limits_{A_i\in ANList}$$Var(A_i)$ & The minimum value in the variances of numeral attributes \\
$f_{21}$ & $\max\limits_{A_i\in ANList}$$Var(A_i)$ & The maximum value in the variances of numeral attributes \\
$f_{22}$ & $Var(Avgs)$, $Avgs$=$\{Avg(A_i)$$|$$A_i\in ANList\}$ & The variance of the average values of the numeral attributes \\
$f_{23}$ & $Var(Vars)$, $Vars$=$\{Var(A_i)$$|$$A_i\in ANList\}$ & The variance of the variances of the numeral attributes \\
\hline
\end{tabular}
}
\label{table3}
\end{table*}

\subsection{The rationality of \textit{Auto-Model}}\label{section:4.1}

We extract the knowledge $InfAll$ from 20 research paper \cite{G3,G4,G11,G15,G18,G1,G2,G5,G6,G7,G8,G9,G10,G12,G13,G14,G16,G17,G19,G20} related to classification algorithms. Considering the classification dataset features in TABLE~\ref{table3} as $Fs$, we construct the inputs of the DMD of \textit{Auto-Model}. Note that since we aim at solving the CASH-Weka problem in the experiments, we only consider the classification algorithms in Weka when generating $InfAll$. Then, we input $InfAll$ and $Fs$ to the AutoModelDMD algorithm, and thus obtain the $KFs=\{f_{1},f_{3},f_{5},f_{7},f_{9},f_{10},f_{13},f_{14},f_{15},f_{16},f_{19}\}$ and $SNA$, a MLP with a suitable architecture, which can select the suitable classification algorithm according to the $KFs$ values of a dataset. In the UDR of \textit{Auto-Model}, for each classification dataset $D$, we input ($D$,$KFs$,$SNA$) to the AutoModelUDR approach and thus get a solution for $D$, i.e., a classification algorithm with a hyperparameter setting. And we can examine the effectiveness of $\textit{Auto-Model}$ approach by analyzing the solutions provided by \textit{Auto-Model} approach.

\begin{table}[t]
\newcommand{\tabincell}[2]{\begin{tabular}{@{}#1@{}}#2\end{tabular}}
\caption{The 50 (Weka) classification algorithms involved in the related paper analyze by our Auto-Model.}\smallskip
\centering
\resizebox{\columnwidth}{!}{
\smallskip\begin{tabular}{|l|l|}
\hline
\textbf{Algorithm Type} & \textbf{Algorithm Name} \\
\hline
weka.classifiers.lazy & IBk, IB1, KStar, LWL \\
\hline
weka.classifiers.meta & \tabincell{l}{AdaBoostM1, AdditiveRegression, Bagging, Decorate, LogitBoost, \\ ClassificationViaRegression, RandomSubSpace, RandomCommittee, \\ ClassificationViaClustering, MultiClassClassifier, RotationForest,\\ MultiBoostAB, StackingC} \\
\hline
weka.classifiers.bayes & \tabincell{l}{AODE, BayesNet, ComplementNaiveBayes, HNB, NaiveBayes,\\ NaiveBayesMultinomial, NaiveBayesSimple, NaiveBayesUpdateable} \\
\hline
weka.classifiers.trees & \tabincell{l}{BFTree, J48, SimpleCart, DecisionStump,FT, Id3, LADTree, LMT,\\ NBTree, RandomForest, RandomTree, REPTree} \\
\hline
weka.classifiers.misc & HyperPipes, VFI \\
\hline
weka.classifiers.rules & JRip, PART, OneR, Ridor, ZeroR \\
\hline
weka.classifiers.functions & \tabincell{l}{Logistic, MultilayerPerceptron, RBFNetwork, SimpleLogistic, SMO,\\ LibSVM} \\
\hline
\end{tabular}
}
\label{table4}
\end{table}

\begin{table}[t]
\newcommand{\tabincell}[2]{\begin{tabular}{@{}#1@{}}#2\end{tabular}}
\caption{Notations and their meanings. Suppose $A$ is a classification algorithm in $CAList$ and $D$ is a classification dataset.}\smallskip
\centering
\resizebox{\columnwidth}{!}{
\smallskip\begin{tabular}{|l|l|}
\hline
\textbf{Notations} & \textbf{Meaning} \\
\hline
$CAList$ & A set of classification algorithms which contain in TABLE~\ref{table4} \\
$P(A,D)$ & \tabincell{l}{The performance of $A$ on $D$. We utilize GA algorithm (timelimit=$10^3$s) to obtain\\ the optimal hyperparameter setting $\lambda$ of $A$, use the 10-fold cross-validation accuracy\\ to calculate $f(\lambda,A,D)$ and consider it as $P(A,D)$.} \\
$Pmax(D)$ & \tabincell{l}{The performance score of $A$, which performs the best among $CAList$ on $D$, on $D$.\\ $Pmax(D)$=$\max\limits_{A\in CAList}$$P(A,D)$} \\
$Pavg(D)$ & The average performance of the algorithms in $CAList$ which can process $D$. \\
$CRelations(D)$ & The classification algorithm which corresponds to $D$ in the obtained $CRelations$ \\
$SNA(D)$ & The optimal classification algorithm $SNA$ selects for $D$ \\
\hline
\end{tabular}
}
\label{table5}
\end{table}

Then we explain the rationality of $\textit{Auto-Model}$ approach by analyzing $CRelations$ and $SNA$. In AutoModelUDR, after selecting an algorithm using $SNA$, the other algorithm and their hyperparameter settings will not be considered as the solution any more, and AutoModelDMD only optimizes the hyperparameters of the selected algorithm to obtain the final solution for the given dataset. This design makes \textit{Auto-Model} effective, but if the algorithm selected by $SNA$ is quite inappropriate, this design will be infeasible. Therefore, reasonable design of $SNA$ is crucial, it has a great influence on the rationality of \textit{Auto-Model} approach. Note that, $CRelations$ is the main criterion to evaluate the quality of the $SNA$'s architecture. If the quality of $CRelations$ is poor, the designed $SNA$ will also be invalid. Therefore, both $CRelations$ and $SNA$ have considerable influence on the rationality of \textit{Auto-Model} method. In this part, we will analyze the quality of the obtained knowledge $CRelations$ (Section~\ref{section:4.1.1}) and the effectiveness of the obtained decision-making model $SNA$ (Section~\ref{section:4.1.2}), and thus explain the rationality of our \textit{Auto-Model}.

TABLE~\ref{table4} shows the all classification algorithms involved in $InfAll$ and TABLE~\ref{table5} gives the notations commonly used in Section~\ref{section:4.1}.

\begin{table*}[t]
\caption{The $SNA(D)$, $PORatio(SNA,D)$, $P(SNA(D),D)$, $Pmax(D)$ and $Pavg(D)$ on different classification datasets used for testing.}\smallskip
\centering
\resizebox{0.95\textwidth}{!}{
\smallskip\begin{tabular}{|c|c|c|c|c|c|c|c|c|c|c|c|c|c|c|c|c|c|c|c|c|c|}
\hline
& D1 & D2 & D3 & D4 & D5 & D6 & D7 & D8 & D9 & D10 \\
\hline
$SNA(D)$ & SimpleCart & RBFNetwork & BayesNet & FT & LibSVM & IBk & FT & IBk & Logistic & SimpleCart  \\
PORatio(SNA,D) & 0.92 & 0.92 & 0.90 & 1.00 & 0.88 & 1.00 & 0.98 & 0.98 & 0.92 & 0.86 \\
P(SNA(D),D) & 0.93 & 0.63 & 0.66 & 0.75 & 0.87 & 0.74 & 0.85 & 0.72 & 0.75 & 0.97 \\
Pmax(D) & 0.99 & 0.94 & 0.77 & 0.75 & 0.99 & 0.74 & 0.97 & 0.95 & 0.89 & 0.97 \\
Pavg(D) & 0.92 & 0.55 & 0.55 & 0.67 & 0.83 & 0.70 & 0.81 & 0.57 & 0.58 & 0.94 \\
\hline
\end{tabular}
}
\label{table8}
\end{table*}

\begin{table*}[t]
\caption{The $SNA(D)$, $PORatio(SNA,D)$, $P(SNA(D),D)$, $Pmax(D)$ and $Pavg(D)$ on different classification datasets used for testing (Continued).}\smallskip
\centering
\resizebox{0.95\textwidth}{!}{
\smallskip\begin{tabular}{|c|c|c|c|c|c|c|c|c|c|c|c|c|c|c|c|c|c|c|c|c|c|}
\hline
& D11 & D12 & D13 & D14 & D15 & D16 & D17 & D18 & D19 & D20 & D21 \\
\hline
$SNA(D)$ & RandomSubSpace & FT & SimpleCart & LWL & RBFNetwork & HNB & J48 & LibSVM & SimpleLogistic & J48 & Logistic \\
PORatio(SNA,D) & 0.82 & 1.00 & 0.82 & 0.54 & 0.80 & 0.84 & 0.92 & 1.00 & 0.88 & 1.00 & 1.00 \\
P(SNA(D),D) & 0.75 & 1.00 & 0.85 & 0.64 & 0.95 & 0.94 & 0.91 & 1.00 & 0.78 & 0.82 & 1.00 \\
Pmax(D) & 0.99 & 1.00 & 0.98 & 0.86 & 0.97 & 0.99 & 0.99 & 1.00 & 1.00 & 0.82 & 1.00 \\
Pavg(D) & 0.68 & 0.95 & 0.84 & 0.59 & 0.93 & 0.83 & 0.69 & 0.98 & 0.67 & 0.79 & 0.84 \\
\hline
\end{tabular}
}
\label{table9}
\end{table*}

\subsubsection{The Quality of Knowledge $CRelations$}\label{section:4.1.1}

In the DMD part of \textit{Auto-Model}, after inputting $InfAll$ to the KnowledgeAcquisition approach (Algorithm~\ref{alg:CA}), $CRelations$=$\{(D_i,OA_{D_i}) | i=1,\ldots ,69\}$, which contains 69 (dataset, best algorithm) pairs, is obtained. The meaning of a pair $(D_i,OA_{D_i})$ in $CRelations$ is as follows: the classification algorithm $OA_{D_i}$ is quite suitable for dealing to the classification dataset $D_i$. If the ability of $OA_{D_i}$ to deal with $D_i$ is better than most of classification algorithms, then this information is valid. And if almost all pairs in $CRelations$ are valid, then we can say that the quality of $CRelations$ is very high. Based on this idea, we design $PORatio$ to quantify the quality of the $CRelations$.

\textbf{Definition 1. (Performance Over Ratio, PORatio)} Consider a classification algorithm $A\in CAList$, and classification dataset $D$ contained in $CRelations$. The Performance Over Ratio ($PORatio$) of $A$ on $D$ is defined as:
\begin{equation}
PORatio(A,D)=\frac{|\{A_i|P(A_i,D)\leq P(A,D)|A_i\in CAList\}|}{|CAList|}
\end{equation}
$PORatio(A,D)$ is the proportion of the algorithms in $CAList$ that are not more effective than $A$ on $D$. It ranges from 0 to 1, and its higher value means the stronger ability of $A$ to solve $D$ and the fewer number of classification algorithms that outperform $A$ on $D$.

$PORatio(CRelations(D),D)$ can measure the validity of a pair $(D,CRelations(D))$ in $CRelations$ effectively. We then can utilize the average $PORatio$ of $CRelations(D)$ over all classification datasets contained in $CRelations$ to quantify the quality of $CRelations$.

\begin{table}[t]
\caption{The average $PORatio$ over all classification datasets in $CRelations$.}\smallskip
\centering
\resizebox{\columnwidth}{!}{
\smallskip\begin{tabular}{|c|c|c|c|c|}
\hline
& \textbf{\textit{CRelations(D)}} & \textbf{Top1-RandomForest } & \textbf{Top2-FT} & \textbf{Top3-RandomTree} \\
\hline
Average $PORatio$ & 0.84 & 0.82 & 0.79 & 0.77 \\
\hline
\end{tabular}
}
\label{table6}
\end{table}

\begin{table}[t]
\caption{The average performance score $P$ over all classification datasets in $CRelations$.}\smallskip
\centering
\resizebox{\columnwidth}{!}{
\smallskip\begin{tabular}{|c|c|c|c|c|}
\hline
& \textbf{\textit{CRelations(D)}} & \textbf{Top1-RandomTree} & \textbf{Top2-REPTree} & \textbf{Top3-J48} \\
\hline
Average $P$ & 0.78 & 0.77 & 0.76 & 0.75 \\
\hline
\end{tabular}
}
\label{table7}
\end{table}

\begin{figure}[t]
\centering
\includegraphics[width=7.5cm, height=4.5cm]{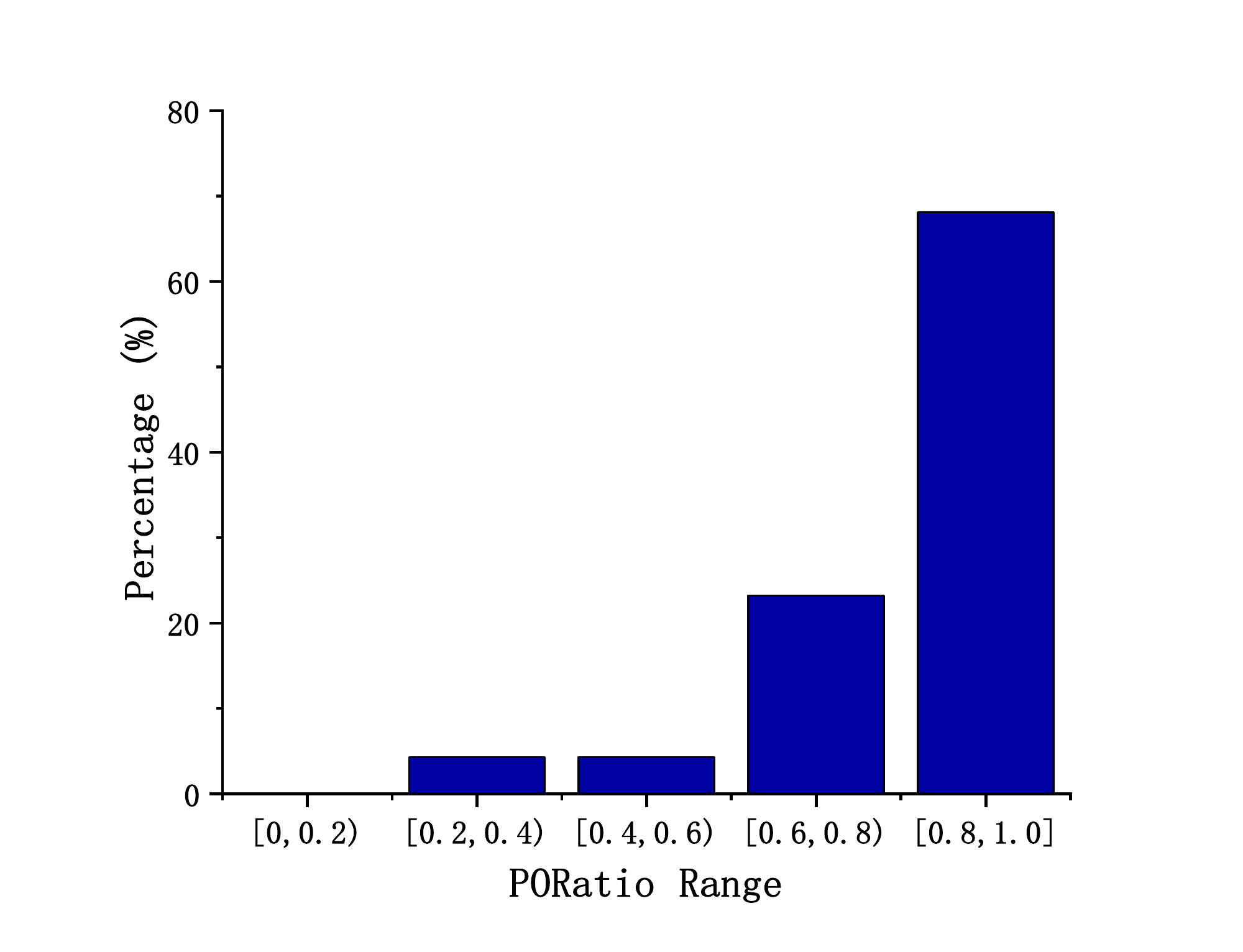}
\caption{The distribution of $PORatio$s of $CRelations(D)$ over all classification datasets in $CRelations$.}
\label{figure3}
\end{figure}

\textbf{Experimental Results.} We calculate the average $PORatio$ of $CRelations(D)$ over all classification datasets in $CRelations$, and analyze the distribution of $PORatio$s of $CRelations(D)$ over all classification datasets in $CRelations$, results are shown in TABLE~\ref{table6} and Fig.~\ref{figure3}. We can observe that, the validity of the pairs in $CRelations$ is generally high, and the quality of the obtained $CRelations$ is high. This shows that the KnowledgeAcquisition approach is effective, and it is feasible to acquire the correspondence between the instance and its optimal algorithm from the related research papers.

Besides, we examine the average $PORatio$ and the average $P$ of a single algorithm $A\in CAList$ over all datasets in $CRelations$, and report the top 3 values and their corresponding classification algorithms, results are shown in TABLE~\ref{table6} and TABLR~\ref{table7}. We can find that, the overall performance of $CRelation(D)$ outperforms a single classification algorithm. This shows that the obtained $CRelation$ is useful. It means that we can achieve higher performance under the guidance of $CRelation$.

\subsubsection{The Effectiveness of Decision-Making Model $SNA$}\label{section:4.1.2}

The target of $SNA$ is to map the classification dataset $D$ to a classification algorithm that is best suited to $D$. If the solution provided by $SNA$, i.e., $SNA(D)$, outperforms most of the classification algorithms, then $SNA$ is effective and its design is reasonable. Thus, we propose to use $PORatio(SNA,D)=PORatio(SNA(D),D)$ to measure the effectiveness of $SNA$ on $D$, and we then can utilize the $PORatio$ of $SNA$ on different classification datasets to examine the effectiveness of $SNA$.

\begin{table*}[t]
\caption{The average $f(T,D)$ on different classification datasets used for testing.}\smallskip
\centering
\resizebox{\textwidth}{!}{
\smallskip\begin{tabular}{|c|c|c|c|c|c|c|c|c|c|c|c|c|c|c|c|c|c|c|c|c|c|c|}
\hline
Time Limit & Method & D1 & D2 & D3 & D4 & D5 & D6 & D7 & D8 & D9 & D10 & D11 & D12 & D13 & D14 & D15 & D15 & D17 & D18 & D19 & D20 & D21 \\
\hline
\multirow{2}{*}{30s} & Auto-Model & \textbf{0.93} & \textbf{0.58} & \textbf{0.66} & \textbf{0.74} & \textbf{0.82} & \textbf{0.74} & \textbf{0.85} & \textbf{0.68} & \textbf{0.62} & \textbf{0.97} & \textbf{0.72} & \textbf{0.99} & \textbf{0.85} & \textbf{0.63} & \textbf{0.94} & 0.94 & \textbf{0.99} & \textbf{1.00} & \textbf{0.73} & \textbf{0.82} & \textbf{1.00}\\
& Auto-Weka & 0.90 & 0.53 & 0.44 & 0.73 & 0.79 & 0.67 & 0.79 & 0.62 & 0.52 & \textbf{0.97} & 0.71 & 0.98 & 0.81 & 0.57 & \textbf{0.94} & \textbf{0.96} & 0.62 & 0.97 & 0.72 & 0.78 & \textbf{1.00} \\
\hline
\multirow{2}{*}{5min} & Auto-Model & \textbf{0.93} & \textbf{0.60} & \textbf{0.66} & \textbf{0.76} & \textbf{0.86} & \textbf{0.74} & \textbf{0.85} & \textbf{0.71} & \textbf{0.73} & \textbf{0.97} & \textbf{0.73} & \textbf{1.00} & \textbf{0.85} & \textbf{0.62} & \textbf{0.94} & \textbf{0.94} & \textbf{0.99} & \textbf{1.00} & \textbf{0.77} & \textbf{0.82} & \textbf{1.00}  \\
& Auto-Weka & 0.89 & 0.49 & 0.44 & 0.72 & 0.83 & 0.69 & 0.77 & 0.62 & 0.54 & 0.96 & 0.71 & \textbf{1.00} & 0.81 & 0.55 & 0.89 & \textbf{0.94} & 0.65 & 0.97 & 0.72 & 0.80 & \textbf{1.00} \\
\hline
\end{tabular}
}
\label{table13}
\end{table*}

\begin{table}[t]
\newcommand{\tabincell}[2]{\begin{tabular}{@{}#1@{}}#2\end{tabular}}
\caption{The 21 classification datasets used for testing. These datasets are not included in $CRelations$}\smallskip
\centering
\resizebox{\columnwidth}{!}{
\smallskip\begin{tabular}{|c|c|c|c|c|c|c|}
\hline
Dataset & Symbol & Records & Attributes & \tabincell{c}{Numeral\\ attributes} & \tabincell{c}{Categorical\\ attributes} & Classes \\
\hline
\tabincell{c}{Pittsburgh Bridges (MATERIAL)} & D1 & 108 & 13 & 3 & 10 & 3 \\
\tabincell{c}{Pittsburgh bridges (TYPE)} & D2 &108 & 13 & 3 & 10 & 6 \\
Flags & D3 & 194 & 30 & 10 & 20 & 8 \\
Liver Disorders & D4 & 345 & 7 & 6 & 1 & 2 \\
Vertebral Column & D5 & 310 & 6 & 5 & 1 & 2 \\
Planning Relax & D6 & 182 & 13 & 12 & 1 & 2 \\
\tabincell{c}{Mammographic Mass} & D7 & 961 & 6 & 1 & 5 & 2 \\
\tabincell{c}{Teaching Assistant Evaluation} & D8 & 151 & 6 & 1 & 5 & 3 \\
Hill-Valley & D9 & 606 & 101 & 100 & 1 & 2 \\
\tabincell{c}{Ozone Level Detection} & D10 & 2536 & 73 & 72 & 1 & 2 \\
Breast Tissue & D11 & 106 & 10 & 9 & 1 & 6 \\
banknote authentication & D12 & 1372 & 5 & 4 & 1 & 2 \\
\tabincell{c}{Thoracic Surgery Data} & D13 & 470 & 17 & 3 & 14 & 2 \\
Leaf & D14 & 340 & 16 & 14 & 2 & 30 \\
\tabincell{c}{Climate Model Simulation Crashes} & D15 & 540 & 19 & 18 & 1 & 2 \\
Nursery & D16 & 12960 & 8 & 0 & 8 & 3 \\
Avila & D17 & 20867 & 10 & 9 & 1 & 12 \\
Chronic\_Kidney\_Disease & D18 & 400 & 25 & 14 & 11 & 2 \\
Crowdsourced Mapping & D19 & 10546 & 29 & 28 & 1 & 6 \\
default of credit card clients & D20 & 30000 & 24 & 14 & 10 & 2 \\
Mice Protein Expression & D21 & 1080 & 82 & 78 & 4 & 8 \\
\hline
\end{tabular}
}
\label{table10}
\end{table}

\begin{table}[t]
\caption{The average $PORatio$ over all classification datasets used for testing.}\smallskip
\centering
\resizebox{\columnwidth}{!}{
\smallskip\begin{tabular}{|c|c|c|c|c|}
\hline
& \textbf{\textit{SNA}} & \textbf{Top1-RandomTree} & \textbf{Top2-FT} & \textbf{Top3-SimpleLogistic} \\
\hline
Average $PORatio$ & 0.90 & 0.83 & 0.83 & 0.78 \\
\hline
\end{tabular}
}
\label{table11}
\end{table}

\begin{table}[t]
\caption{The average performance score $P$ over all classification datasets in $CRelations$.}\smallskip
\centering
\resizebox{\columnwidth}{!}{
\smallskip\begin{tabular}{|c|c|c|c|c|}
\hline
& \textbf{\textit{SNA(D)}} & \textbf{Top1-RandomTree} & \textbf{Top2-RepTree} & \textbf{Top3-NaiveBayes} \\
\hline
Average $P$ & 0.83 & 0.81 & 0.80 & 0.79 \\
\hline
\end{tabular}
}
\label{table12}
\end{table}

\textbf{Experimental Results.} We record the $SNA(D)$ and calculate the $PORatio(SNA,D)$, $P(SNA(D),D)$, $Pmax(D)$ and $Pavg(D)$ on different classification datasets in TABLE~\ref{table10}, results are shown in TABLE~\ref{table8} and TABLE~\ref{table9}. We can observe that, the $PORatio(SNA,D)$ is generally very high, and $P(SNA(D),D)$ is always superior to $Pavg(D)$. This shows that the $SNA$ designed by the DMD part of \textit{Auto-Model} is reasonable and effective, and the design of AutoModelUDR approach is feasible.

Besides, we examine the average $PORatio$ and the average $P$ of a single algorithm $A\in CAList$ over the classification datasets in TABLE~\ref{table10}, and report the top 3 values and their corresponding classification algorithms, results are shown in TABLE~\ref{table11} and TABLR~\ref{table12}. We can find that, the overall performance of $SNA(D)$ outperforms a single classification algorithm. This shows that the obtained $SNA$ is effective, i.e. $SNA$ can select quite appropriate algorithm, and thus help us achieve better performance. Two key contents of \textit{Auto-Model} approach, i.e., $CRelations$ and $SNA$, are proved to be reasonable and effective. Therefore, the whole design of \textit{Auto-Model} approach is feasible and rational.

\subsection{Compare \textit{Auto-Model} with Auto-Weka}\label{section:4.2}

In this part, we examine the ability of \textit{Auto-Model} approach and Auto-Weka approach to deal with the CASH-Weka problem, and thus compare their effectiveness. Notations that are commonly used in this section are shown in TABLE~\ref{table15}.

For each classification dataset $D$ used for testing, we divide it into 10 folds equally and utilize the measure $f(T,D)$ defined in Table~\ref{table15}, where $T$ is \textit{Auto-Model} or Auto-Weka, to examine the effectiveness of $T$. The higher $f(T,D)$ is, the better the solution $T(D)$ is and thus the more effective $T$ is. We also analyze the effectiveness of Auto-Weka and \textit{Auto-Model} under different time limits, results are shown in TABLE~\ref{table13}. Note that, for each $f(T,D)$, we calculate it 20 times, and report the average value in TABLE~\ref{table13}. We can observe that \textit{Auto-Model} can often obtain better solutions within short time (30 minutes), and the quality of the solutions provided by it improves more markedly when the time limit becomes longer (5 minutes).

Let us analyze the reasons. \textit{Auto-Model} can efficiently select a quite suitable classification algorithm with the help of reasonable designed $SNA$, and utilize the left time to find the optimal hyperparameter setting for optimizing the performance of selected algorithm, whereas, Auto-Weka considers a huge search space which contains the algorithms and their hyperparameters, and unable to find out suitable algorithms in a short time. As the comparison, Auto-Weka needs to waste much time on evaluating inappropriate classification algorithms with various hyperparameter settings. Therefore, its performance is lower than that of \textit{Auto-Model}.

Overall, the design of our \textit{Auto-Model} approach is reasonable. \textit{Auto-Model} can provide high-quality solutions for users within shorter time, and tremendously reduces the cost of algorithm implementations. It outperforms Auto-Weka and can more effectively deal with the CASH problem.

\begin{table}[t]
\newcommand{\tabincell}[2]{\begin{tabular}{@{}#1@{}}#2\end{tabular}}
\caption{Notations and their meanings. Suppose $T$ is a CASH technique and in $D$ is a classification dataset.}\smallskip
\centering
\resizebox{\columnwidth}{!}{
\smallskip\begin{tabular}{|l|l|}
\hline
\textbf{Notations} & \textbf{Meaning} \\
\hline
$T(D)$ & \tabincell{l}{The optimal algorithm with the optimal hyperparameter setting\\ provided by $T$ for solving $D$} \\
$f(T,D)$ & \tabincell{l}{The performance of $T(D)$ on $D$. We use the 10-fold cross-validation\\ accuracy to calculate $f(T,D)$.} \\
\hline
\end{tabular}
}
\label{table15}
\end{table}

\section{Conclusion and Future Works}\label{section:5}

In this paper, we propose the \textit{Auto-Model} approach, which makes full use of known information in the research papers and introduces hyperparameter optimization techniques, to help users to effectively select the suitable algorithm and hyperparameter setting for the given problem instance. \textit{Auto-Model} tremendously reduces the cost of algorithm implementations and hyperparameter configuration space, and thus capable of dealing with the CASH problem efficiently and easily. We also design a series of experiments to analyze the reliability of information derived from research papers by our proposed \textit{Auto-Model}, and examine the performance of \textit{Auto-Model} and compare with that of classical Auto-Weka approach. The experimental results demonstrate that the information extracted is relatively reliable, and our \textit{Auto-Model} is more effective and practical than Auto-Weka. In the future works, we will try to design an algorithm to accurately and automatically extract the information we need from the research papers, and thus achieve the total automation of our \textit{Auto-Model} approach. Besides, we tend to utilize our CASH technique to help users to deal with more problems, and develop a system with high usability.

\bibliographystyle{IEEEtran}
\bibliography{ref}

\end{document}